\def\year{2019}\relax
\documentclass[letterpaper]{article} %DO NOT CHANGE THIS
\usepackage{aaai19}  %Required
\usepackage{times}  %Required
\usepackage{helvet}  %Required
\usepackage{courier}  %Required
\usepackage{url}  %Required
\usepackage{graphicx}  %Required
\usepackage{enumitem}
\usepackage{floatrow}
\newfloatcommand{capbtabbox}{table}[][\FBwidth]

\usepackage{booktabs}       % professional-quality tables
\usepackage{amsfonts}       % blackboard math symbols
\usepackage{nicefrac}       % compact symbols for 1/2, etc.

\usepackage[font=small]{subcaption}
\usepackage[font=small]{caption}
\usepackage{algorithm}
\usepackage{algorithmic}
\usepackage{amsthm}
\usepackage{amssymb}
\usepackage{multicol}
\usepackage{amsmath}

\newtheorem{thm}{Theorem}
\newtheorem{lem}{Lemma}

\newtheorem{definition}{Definition}

\usepackage{color}
\definecolor{blue}{RGB}{0, 93, 170}			%Go Big Blue!

\usepackage[textsize=scriptsize]{todonotes}
\newcommand{\citet}[1]{\citeauthor{#1}~\shortcite{#1}}
\newcommand{\citep}{\cite}

\newcommand{\blend}{\ensuremath{\sigma_{online}}}

\nocopyright

\begin{document}

\setlength\titlebox{2.5in}

\title{Incorporating Behavioral Constraints in Online AI Systems}
\author{Avinash Balakrishnan, Djallel Bouneffouf, Nicholas Mattei, Francesca Rossi\thanks{On leave from University of Padova.} \vspace{0.5em} \\
MIT-IBM Watson AI Lab \\
IBM Research \\
Yorktown Heights, NY, USA \\
\small{avinash.bala@us.ibm.com, djallel.bouneffouf@ibm.com, n.mattei@ibm.com, francesca.rossi2@ibm.com}
}

\maketitle
\begin{abstract}
AI systems that learn through reward feedback about the actions they take are increasingly deployed in domains that have significant impact on our daily life. However, in many cases the online rewards should not be the only guiding criteria, as there are additional constraints and/or priorities imposed by regulations, values, preferences, or ethical principles. We detail a novel online agent that learns a set of behavioral constraints by observation and uses these learned constraints as a guide when making decisions in an online setting while still being reactive to reward feedback. To define this agent, we propose to adopt a novel extension to the classical contextual multi-armed bandit setting and we provide a new algorithm called Behavior Constrained Thompson Sampling (BCTS) that allows for online learning while obeying exogenous constraints. Our agent learns a constrained policy that implements the observed behavioral constraints demonstrated by a teacher agent, and then uses this constrained policy to guide the reward-based online exploration and exploitation. We characterize the upper bound on the expected regret of the contextual bandit algorithm that underlies our agent and provide a case study with real world data in two application domains.  Our experiments show that the designed agent is able to act within the set of behavior constraints without significantly degrading its overall reward performance.
\end{abstract}

\section{Introduction}

In online decision settings, an agent must select one out of several possible actions, e.g., recommending a movie to a particular user, or proposing a treatment to a patient in a clinical trial. Each of these actions is associated with a context, e.g., a user profile, and a feedback signal, e.g., the reward or rating, is only observed for the chosen option. In these online decision settings the agent must learn the inherent trade-off between exploration, which involves identifying and understanding the reward from an action, and exploitation, which means gathering as much reward as possible from an action.  We consider cases where the behavior of the online agent may need to be restricted in its choice of an action for a given context by laws, values, preferences, or ethical principles \cite{russell2015research}.  More precisely, we apply a set of \emph{behavioral constraints} to the agent that are independent of the reward function. For instance, a parent or guardian group may want a movie recommender system (the agen)t to not recommend certain types of movies to children, even if the recommendation of such movies could lead to a high reward \cite{BaBoMaRo18}. In clinical settings, a doctor may want its diagnosis support system to not recommend a drug that typically works because of considerations related to the patients' quality of life. 

Many decision problems where an agent is responsive to online feedback are modeled as a {\em multi-armed bandit (MAB)} problem \cite{MaryGP15,villar2015multi}.  In the MAB setting there are $K$ \emph{arms}, each associated with a fixed but unknown reward probability distribution \cite{LR85,UCB}. At each time step, an agent plays an arm, i.e., recommends an item to a user, and receives a reward that follows the selected arm's probability distribution, independent of the previous actions. A popular generalization of MAB is the contextual multi-armed bandit (CMAB) problem where the agent observes a $d$-dimensional {\em feature vector}, or {\em context}, to use along with the rewards of the arms played in the past in order to choose an arm to play. Over time, the agent learns the relationship between contexts and rewards and select the best arm \cite{AgrawalG13}. 

In giving the agent a set of behavioral constraints, we consider the case where only \emph{examples} of the correct behaviors are given by a teacher agent, and our online agent must learn and respect these constraints in the later phases of decision making: As an example, a parent may give examples of movies that their children can watch (or that they cannot watch) when setting up a new movie account for them. 

The idea of teaching machines right from wrong has become an important research topic in both AI \cite{yu2018building} and in other disciplines \cite{wallach2008moral}. Much of the research at the intersection of artificial intelligence and ethics falls under the heading of \emph{machine ethics}, i.e., adding ethics and/or constraints to a particular system's decision making process \cite{anderson2011machine}. One very important task in machine ethics is \emph{value alignment}, i.e., the idea that an agent can only pursue goals that are aligned to human values and are therefore beneficial to humans \cite{russell2015research,LoMaRoVe18,LoMaRoVe18a}. However, this still leaves open the question of how to provide the values, or the behavioral constraints derived from the values, to the agent. 
A popular technique is called the \emph{bottom up approach}, i.e., teaching a machine what is right and wrong by example \cite{allen2005artificial}. We adopt this technique in our paper, since we consider the case where only \emph{examples} of both correct and incorrect behavior are given to the agent, that must learn from these.  
%Within the RL community this can be seen as a particular type of apprenticeship learning \cite{abbeel2004apprenticeship} where the agent is learning how to be \emph{safe}, rather than only maximizing reward \cite{leike2017ai}, by exploiting both positive and negative examples of safe behavior.
Note that having only examples of good and bad  behavior means that we need to learn the constraints from such examples. On the contrary, if the constraints were known and we received feedback on them during the recommendation phase, then we could use the stochastic combinatorial semi-bandit setting \cite{matroid14}. Also, if the constraints were known and budget-like, then we could use bandits with knapsacks \cite{AgDe16a}.  However, the case where we are only given examples of the behavioral constraints to guide the online agent's  behavior does not seem to be covered by the current literature.

%These exogenous behavioral constraints are guidelines that should be followed by the agent and not updated online by personalization or modification. For this reason, they should be fixed, either explicitly given or learnt from examples, before the online agent begins taking actions. In many settings we may not have access to the constraints as an explicit set of rules or function, we may only have access to examples of constrained behavior from a doctor, or a set of decisions made by a parent.  In this paper we start with examples of constrained behavior and want to build online agents who are able to learn and follow these constraints.

To give flexibility to our agent, we let the system designer to decide how much the guidelines given by the behavioral constraints should weigh on the decision of the agent during the online phase. So, to control the tradeoff between following the learned behavioral constraints and pursuing a greedy online-only policy, we expose a parameter of the algorithm called \blend.  This parameter allows the system designer to smoothly transition between the two policy extremes, where $\blend = 0.0$ means that we are only following the learned constraints and are insensitive to the online reward, while $\blend= 1.0$ means we are only following the online rewards and not giving any weight to the learned constraints.

\smallskip
\noindent
\textbf{Contributions.} We propose a novel extension of the contextual bandit setting that we call the \emph{Behavior Constrained Contextual Bandits Problem (BCCBP)}, where the  agent is constrained by a policy that it has learned from observing examples of good and bad behavior. % that it then emulates during the recommendation phase. 
We provide a new algorithm for this setting, that we call \emph{Behavior Constrained Thompson Sampling (BCTS)}, that is able to trade-off between the constrained behavior learned from examples given by a teaching agent and reward-driven behavior learned during online recommendation. We prove an upper bound on the expected regret of this new algorithm and evaluate it empirically on data from two real world settings. The experimental evaluation shows that it is possible to learn and act in a constrained manner while not significantly degrading the performance.

%In this two stage algorithm the CTS agent first learns a behavior constrained policy from observations and then uses this policy to guide its actions in the online setting. The main contributions of this paper are (1) introducing a new formulation of a contextual bandit problem, motivated by practical applications with constrained decision, (2) proposing a new algorithm, which extend the existing bandit algorithms to the new setting, (3) proving an upper bound on the expected regret of our algorithm, and (4) evaluating the algorithm empirically on a movie recommendation problem. The experimental evaluation shows that it is possible to learn and act in a constrained manner while not significantly degrading the performance of the online agent.

\section{Background and Related Work}
\label{sec:related}

\smallskip
\noindent
\textbf{Multi-armed Bandit Settings.} The classic multi-armed bandit (MAB) problem is a model of the trade-off between exploration and exploitation, where an agent acting in a live, online environment wants to pick, within a finite set of decisions, the one maximizing the cumulative reward \cite{LR85,UCB}.  The MAB problem is a classic example of reinforcement learning, where the online agent receives signals that are then used to update their behavior.  Reinforcement learning, and the MAB problem specifically, have been used to successfully solve a number of real-world problems \cite{SuBa17a,Mnih2013}.

The contextual multi-armed bandit (CMAB) problem, a generalization of the MAB problem that exploits the presence of features for users of the online decision system, has been studied in multiple domains including recommender systems and online advertising. Optimal solutions have been provided by using a stochastic formulation ~\cite {LR85,UCB}, a Bayesian formulation ~\cite {T33,kaufmann,AgrawalG12}, and an adversarial formulation ~\cite{AuerC98,AuerCFS02}. 
%However, these approaches do not take into account the context which may affect the arm's performance. 
Both LINUCB \cite{Li2010,ChuLRS11} and Contextual Thompson Sampling (CTS)\cite{AgrawalG13} assume a linear dependency between the expected reward of an action and its context; the representation space is modeled using a set of linear predictors and provide bounds on the expected regret.

The literature shows some recent work on combining the contextual bandit formalism with constraints. \cite{WuSLJ15} propose a contextual bandits with budget and time constraints, that are expressed over the number of times, and time period, that an arm can be pulled. Such coupling effects make it difficult to obtain oracle solutions that assume known statistics of bandits. The authors develop an approximation of the oracle, referred to as Adaptive-Linear-Programming (ALP), which achieves near-optimality and only requires the ordering of expected rewards. 

Additionally, \citet{AgDe16a} consider multi-faceted budget constraints where each arm pull exhausts some facet of a budget that needs to be optimized under. \citet{BouneffoufRCF17} consider a novel formulation of the contextual bandit problem when there are constraints on the context, i.e., where only a limited number of features can be accessed by the learner at each iteration. This novel formulation is motivated by different online problems arising in clinical trials and recommender systems where accessing all parts of the context could be costly. %They propose a novel algorithm, called Thompson Sampling with Restricted Context (TSRC) for handling stationary bandits with restricted contexts.  
None of these formalisms capture the setting we consider in this paper, as the budgets or time constraints are explicitly supplied a priori.

There are also a number of closely related bandit formalisms including matroid bandits, also called stochastic combinatorial semi-bandits, contextual bandits with history, and conservative bandits. \cite{matroid14} consider matroid bandits which are able to optimize combinatorial functions that are expressible as matroids, e.g., linear combinations of objectives.  In relation to our work, if the constraints were known to the agent, then we could leverage these algorithms.  
However, we assume that our agent does not receive the constraints explicitly during the constraint learning phase and does not receive feedback on the constraints during the recommendation phase. We believe that in many settings we may not have access to the constraints as an explicit set of rules or a function, rather we may only have access to (positive and negative) examples of constrained behavior from a doctor, or a set of (good and bad) decisions made by a parent, and that this "teaching" figure may not always be able to provide feedback on each recommendation. 

\citet{history12} introduces the setting called contextual bandit with history: an agent is furnished with examples of past behavior over the same reward function and leverages these observations to guide its future behavior. In our setting we instead have two separate reward functions, the constraints and the online reward, and hence the results are not directly applicable.  

Finally, \citet{wu2016conservative} introduces conservative bandits, which attempt to maximize the achieved reward of an online agent while keeping the cumulative reward above a certain threshold. These setting is useful, for example, when one wants to try out new advertisements but does not want revenue to fall below a certain threshold. This work is fundamentally different from our own as we want the agent to not exploit certain arms at all rather than maintain a minimum of overall reward.

\smallskip
\noindent
\textbf{Constrained and Ethical Decision Making.\;} Humans often constrain the decisions that they take according to a number of exogenous priorities, derived by moral, ethical, religious, or business values \cite{Sen}.  Constrained or ethical decision making has been studied in a variety of contexts in computer science with most of the work focused on teaching a computer system to act within guidelines \cite{bonnefon2016social}.  Broadly, our work fits into constrained reinforcement learning / Safe RL \cite{leike2017ai} and value alignment \cite{russell2015research,LoMaRoVe18,LoMaRoVe18a}.

For example, \citet{briggs2015sorry} discusses a rule-based system applied to scenarios in which a robot should infer that a directive leads to undesirable behavior. In this system, given some instructions, the agent first reasons about a set of conditions including ``Do I know how to accomplish the task?'', and ``Does accomplishing this task violate any normative principles?''. Each of these conditions is formalized as a logical expression, along with inference rules that enable the agent to infer which directives to reject.

Having the agent learn about its ethical objective function while making decisions based on this objective function is a challenging problem. In \cite{armstrong2015motivated} the authors consider this problem by exploring the consequences of an agent that uses Bayesian learning to update its beliefs about the ``true'' ethical objective function. At each time step, the agent makes decisions that maximize a utility function based on the agents beliefs about the ethical objective function.
In reinforcement learning, Markov decision processes have been used to study both ethics and ethical decision making.  In \cite{abel2016reinforcement} the authors argue that the reinforcement learning framework achieves the appropriate generality required to theorize about an idealized ethical artificial agent.  

%They claim that the RL framework offers the proper foundations for grounding specific questions about ethical learning and decision making and they propose the RL framework be use for further scientific investigation. To support this point they define an idealized formalism for an ethical learner and conduct experiments to show the soundness of their approach.

None of these previously proposed approaches to ethical or behavior-constrained decision making leverage the contextual bandit setting.  We instead feel that the bandit setting, which has broad application across computer science and decision making, is an ideal formalism for online decision making and can be fruitfully extended to include behavioral constraints.

\section{Preliminaries}

Following \citet{11}, the contextual bandit problem is defined as follows.
At each time $t \in \{1,...,T\}$, a player is presented with a {\em context vector} $c(t) \in \mathbf{R}^d$ and must choose an arm $k  \in K = \{ 1,...,|K|\}$.
%We will denote by $C=\{C_1,...,C_N\}$  the set of features (variables) defining the context.
%
Let ${\bf r} = (r_{1}(t),...,$ $r_{K}(t))$ denote  a reward vector, where $r_k(t) \in [0,1]$ is a reward at time $t$ associated with the arm $k\in A$.
%Let $\pi: C \rightarrow A$ denote a policy and $D_{c,r}$ denote a joint distribution  $({\bf c},{\bf r})$.
%
We assume that the expected reward is a linear function of the context, i.e. $E[r_k(t)|c(t)] = \mu_k^T c(t)$, where $\mu_k$ is an unknown weight vector (to be learned from the data) associated with arm $k$.

The purpose of a contextual bandit algorithm $A$ is to minimize the cumulative regret. Let $H:C\rightarrow [K]$ where $C$ is the set of possible contexts and $c(t)$ is the context at time $t$, $h_t \in H$ a hypothesis computed by the algorithm $A$ at time $t$ and $h^{*}_{t}=\underset{h_t \in H}{\operatorname{argmax}}\ r_{h_{t}(c(t))}(t)$ the optimal hypothesis at the same round. The cumulative regret is: $R(T) = \sum ^{T}_{t=1} {r_{h^{*}_{t}(c(t))}(t)- r_{h_t(c(t))}(t)}$.

One widely used way to solve the contextual bandit problem is the Contextual Thompson Sampling algorithm (CTS) \cite{AgrawalG13} given as Algorithm \ref{alg:CTS}.
In CTS, the reward $r_{k}(t)$ for choosing arm $k$ at time $t$ follows a parametric likelihood function $Pr(r(t)|\tilde{\mu})$.  
Following \cite{AgrawalG13}, the posterior distribution at time $t + 1$, $Pr(\tilde{\mu}|r(t)) \propto Pr(r(t)|\tilde{\mu}) Pr(\tilde{\mu})$ is given by a multivariate Gaussian distribution $\mathcal{N}(\hat{\mu_k}(t+1)$, $v^2 B_k(t + 1)^{-1})$, where
$B_k(t)= I_d + \sum^{t-1}_{\tau=1} c(\tau) c(\tau)^\top$, $d$ is the size of the context vectors $c$, $v= R \sqrt{\frac{24}{z} d \cdot ln(\frac{1}{\gamma})}$ and we have $R>0$,  $z \in [0,1]$, $\gamma \in [0,1]$ constants, and $\hat{\mu}(t)=B_k(t)^{-1} (\sum^{t-1}_{\tau=1} c(\tau) r_k(\tau))$.

Every step $t$ consists of generating a $d$-dimensional sample $\tilde{\mu_k}(t)$ from $\mathcal{N}(\hat{\mu_k}(t)$, $ v^2B_k(t)^{-1})$ for each arm.  We then decide which arm $k$ to pull by solving for $argmax_{k\in K} c(t)^\top \tilde{\mu_k}(t)$.  This means that at each time step we are selecting the arm that we expect to maximize the observed reward given a sample of our current beliefs over the distribution of rewards, $c(t)^\top \tilde{\mu_k}(t)$.  We then observe the actual reward of pulling arm $k$, $r_k(t)$ and update our beliefs.
\begin{definition}[Optimal Policy]
 The optimal policy for solving the contextual MAB is selecting the arms at time $t$ :
%\begin{align*}
$k(t) = \underset{k\in K }{argmax} \;\tilde{\mu}_{k}^*(t)^\top c(t)$, 
%\end{align*}
%\noindent
where $\tilde{\mu}_{k}^*$ the optimal mean vector for the reward driven policy.
\end{definition}

\begin{algorithm}
   \caption{Contextual Thompson Sampling Algorithm}
%   \label{alg.2}
\label{alg:CTS}
\begin{algorithmic}[1]
 \STATE {\bfseries }\textbf{Initialize:}  $B= I_d$, $ \hat{\mu}= 0_d, f = 0_d$.
 \STATE {\bfseries }\textbf{Foreach} $t = 1, 2, . . . ,T$ \textbf{do}
 \STATE {\bfseries }\quad Sample $\tilde{\mu_{k}}(t)$ from the $N(\hat{\mu}_k, v^2 B_k^{-1})$ distribution.
 \STATE {\bfseries }\quad Play arm $k_t= \underset{k\in K}{argmax}\ c(t)^\top \tilde{\mu_{k}}(t) $
% \STATE {\bfseries } \textbf{else if} not all $x \in c$ are queried \textbf{goto} 5.
  \STATE {\bfseries }\quad Observe $r_{k}(t)$ 
 \STATE  $B_k= B_{k}+ c(t)c(t)^{T} $, $f = f + c(t)r_{k}(t)$, $\hat{\mu_k} = B_k^{-1} f$
 \STATE {\bfseries }\textbf{End}
   \end{algorithmic}
\end{algorithm}

\section{Behavior Constrained Contextual Bandits}
\label{subsec:UCB}
Here we define a new type of a bandit problem, the {\em Behavior Constrained Contextual Bandits (BCCB)}, present a novel algorithm and agent for solving this problem, and derive an upper bound for both the expected online regret and the regret as a function of the number of constraint examples given by the teacher.  

In this setting, the agent first goes through a {\em constraint learning phase} where it is allowed to query the user $N$ times and receive feedback $r^e_k(t) \in [0,1]$ about whether or not the chosen decision is allowed under the desired constrained behavior during the recommendation phase. During the {\em online recommendation phase}, the goal of the agent is to maximize both $r_k(t) \in [0,1]$, the reward of the action $k$ at time $t$, while minimizing the (unobserved) $r^e_k(t) \in [0,1]$, which models whether or not the pulling of arm $k$ violates the behavioral constraints. During the recommendation phase, the agent receives no feedback on the value of $r^e_k (t)$, as the labeler may not be around to always provide this feedback. In order to follow and subsequently maximize alignment with the behavioral constraints the agent must learn an effective policy that implements the constraints during the first phase and apply it during the recommendation phase.  In the second phase we are interested in the total \emph{behavioral error} incurred by the agent, i.e., $E(T) = \sum ^{T}_{t=1} r^e_{h_t(c(t))}(t)$.

\subsection{Behavior Constrained Thompson Sampling}

Our agent employs an extension of the classical Thompson sampling algorithm to first learn a constrained policy from observation and then use this constrained policy to guide the online learning task. Behavior Constrained Thompson Sampling (BCTS) contains two parts:

\begin{itemize}
\item \textbf{Constraint Learning Phase}:
During this phase the agent learns the behavioral constraints through interaction with a teaching agent that is able to, at each iteration, provide a context to our agent as well as feedback as to whether or not the action chosen by our agent is allowed.  This feedback takes the form of a binary reward revealed for the arm $k$ that is chosen at time $t$, $r^e_k(t) \in [0,1]$. We use the classical Thompson sampling (CTS) algorithm (see Algorithm \ref{alg:ETS} line 2) in order to explore the constraint space and learn a policy.  We also compare this against the setting where the agent chooses random arms during the teaching phase. We call the behavior that the agent learns during this phase the constrained policy $\mu^e$ where $\hat{\mu}^e_k$ and $\hat{B}_k$ denotes respectively the learned mean vector and the covariance matrix for each arm $k$.

\item \textbf{Online Recommendation Phase}: 
During the recommendation phase the agent continually faces a dilemma: follow the constrained policy or follow the reward signal. Our algorithm uses the Thompson sampling strategy to estimate the expected rewards of the online policy for each arm (Algorithm \ref{alg:ETS} line 4--7), while using the constrained policy to estimate the expected behavior $\tilde{\mu_k}(t)$. It then computes a weighted combination of $\tilde{\mu}^e_k(t) $ and $\tilde{\mu_k}(t)$ for each arm using $\blend$ as weight given by the user (line 15), this weight balances between following a reward driven policy and constrained policy. It then obtains the reward (line 16) and updates the parameters of the distribution for each $\hat{\mu_k}$ (line 17). Finally, the reward $r_k(t)$ for the chosen arm $k$ is observed, and relevant parameters are updated.
\end{itemize}

\begin{algorithm}[t]
 \caption{Behavior Constrained Thompson Sampling}
% \label{alg.2}
\label{alg:ETS}
\begin{algorithmic}[1]
 %$C^d$ denotes an arbitrary subset of $C$ of size $d$.
 \STATE {\bfseries }\textbf{Initialize:} $\forall k \in K, B_k=I_d$, $V_{\tilde{\mu}}=0_d$, $V_{\mu^e}=0_d$,
 $\hat{\mu_k}= 0_d, g_k = 0_d$, $\blend$.
\STATE {\bfseries }\textbf{// Constraint Learning Phase}
\STATE {\bfseries }\textbf{Foreach} $t = 1, 2, . . . ,N$ \textbf{do}
		\STATE {\bfseries } \quad $c(t)$ is revealed to the agent.
		\STATE {\bfseries }  \quad The agent chooses an action $k$. %~~~ // We use CTS (Alg. 1) to learn a constrained policy 
		\STATE {\bfseries } \quad The teaching agent reveals reward $r^e_k(t)$.
		\STATE {\bfseries } \quad The agent updates its policy $\mu^e_t$. 
		\STATE {\bfseries } \quad $t=t+1$
%\STATE {\bfseries } \quad\textbf{End do}
\STATE {\bfseries }\textbf{// Online Recommendation Phase}
 \STATE {\bfseries }\textbf{Foreach} $t = 1, 2, . . . ,T$ \textbf{do}
 \STATE  {\bfseries }\quad
 Observe the context vector $c(t)$ of features.
 \STATE {\bfseries }\quad\textbf{Foreach} arm $k \in K$ \textbf{do}
 \STATE {\bfseries }\quad \quad Sample $\tilde{\mu_k}(t)$ from $N(\hat{\mu_k}, v^2 B_k^{-1})$ distribution.
  \STATE {\bfseries }\quad \quad Sample $\tilde{\mu}^e_k(t)$ from $N(\hat{\mu}^e_k, v^2 \hat{B}_k^{-1})$ distribution.
 \STATE {\bfseries } \quad\textbf{End do}
 \STATE {\bfseries } \quad Select $k(t)=  \underset{k\in K }{argmax} ~\blend \cdot \tilde{\mu_k}(t)^\top c(t) + (1-\blend) \tilde{\mu}^e_k(t) ^\top c(t)$
 \STATE {\bfseries }\quad Observe $r_{k}(t)$
\STATE \quad $B_k= B_{k}+ c(t)c(t)^{T} $
\STATE \quad $g_k = g_k + c(t)r_{k}(t)$
\STATE \quad $\hat{\mu_k} = B_k^{-1} g_k$
%\STATE \quad $g_k = g_k + c(t)r_{k}(t)$
%\STATE \quad $\hat{\mu_k} = B_k^{-1} g_k$
 \STATE {\bfseries }\textbf{End do}
 \end{algorithmic}
\end{algorithm}

We derive an upper bound on the regret $R(T)$ of the policy computed by BCTS. We use the standard definition of the optimal policy from the Contextual MAB literature. Note that this definition is used for most bandit problems.  We are also interested in studying the the behavioral error $E(T)$ incurred by the agent for any policy it may follow in the online phase.

\begin{thm}\label{thm:upperbcts} 
Consider the BCTS algorithm with $K$ arms, $d$ features, and take  $0 \leq \blend \leq 1$ and $0 \leq \gamma \leq 1$. % and $\blend$ the distance 
Then, with probability $(1-\gamma)$, the upper bound on the regret R(T) at time $T$  %\ref{alg:ETS}) 
is:
\begin{eqnarray*}
%E(R(T))<
\blend \frac{d\gamma}{z} \sqrt{T^{z+1}} (ln (T) d) ln \frac{1}{\gamma}+ \nonumber \\
(1-\blend) c_{max} T ||\tilde{\mu}^*_{max}+\mu^{e}_{min}||_2
\end{eqnarray*}
\noindent
where $c_{max}$ a positive constant and $0<z<1$, a constant parameter of the CTS algorithm, $\blend$ a distance threshold and $\mu^{e}_{max}= max_k(||\mu^{e}_k||_1)$, $\tilde{\mu}^*_{max}= max_k(||\tilde{\mu}^*_k||_1)$ with $k \in K$.

\end{thm}
The above theorem tells us that, if the constrained policy $\mu^{e}$ and the optimal policy $\tilde{\mu}^*$ are close to each other, then we can recover the bound on CTS given as a Lemma by \citet{AgrawalG13}.  Hence, it is interesting for future work to study distributions of the data and find data driven bounds on the regret. 

\begin{definition}[Optimal Policy for BCCB]
 The optimal policy for solving the BCCB is selecting the arm at time $t$:
%\begin{align*}
$k(t) = \underset{k\in K }{argmax} \;(\blend^*\tilde{\mu}^*_{k}(t)+(1-\blend^*)\tilde{\mu}_{k}^{e*}(t))^\top c(t)$, 
%\end{align*}
%\noindent
where $\tilde{\mu}_{k}^*$ the optimal mean vector for the reward driven policy $\tilde{\mu}_{k}^{e*}$ the optimal mean vector for the behavior constraint driven policy.
\end{definition} 

\begin{thm}\label{thm:upperCBTS} 
Using Definition 2 of optimal policy we have, with probability $(1-\gamma)$, the upper bound on the regret R(T) at time $T$ %\ref{alg:ETS}) 
is:
\begin{eqnarray*}
%E(R(T))<
  max(\blend^*,\blend)\frac{d\gamma}{z} \sqrt{T^{z+1}} (ln (T) d) ln \frac{1}{\gamma}+\\max((1-\blend^*),(1-\blend))\frac{d\gamma}{z} \sqrt{N^{z+1}} (ln (N) d) ln \frac{1}{\gamma}
\end{eqnarray*}
\noindent
where $c_{max}$ a positive constant and $0<z<1$, a constant parameter of the CTS algorithm, $\blend^*$ the optimal  distance threshold.

\end{thm}

Theorem \ref{thm:upperCBTS} is interesting in that, if we use Definition 2 as our optimal policy, the regret of our proposed solution is sub-linear in time $T$ and sub-linear in the number of training examples $N$.  We will compare using the CTS algorithm to a random baseline in our experiments in the next section.

\section{Experimental Evaluation}

\begin{figure*}[ht]
\centering
\begin{subfigure}{0.33\linewidth}
	\centering
	\includegraphics[width=\linewidth]{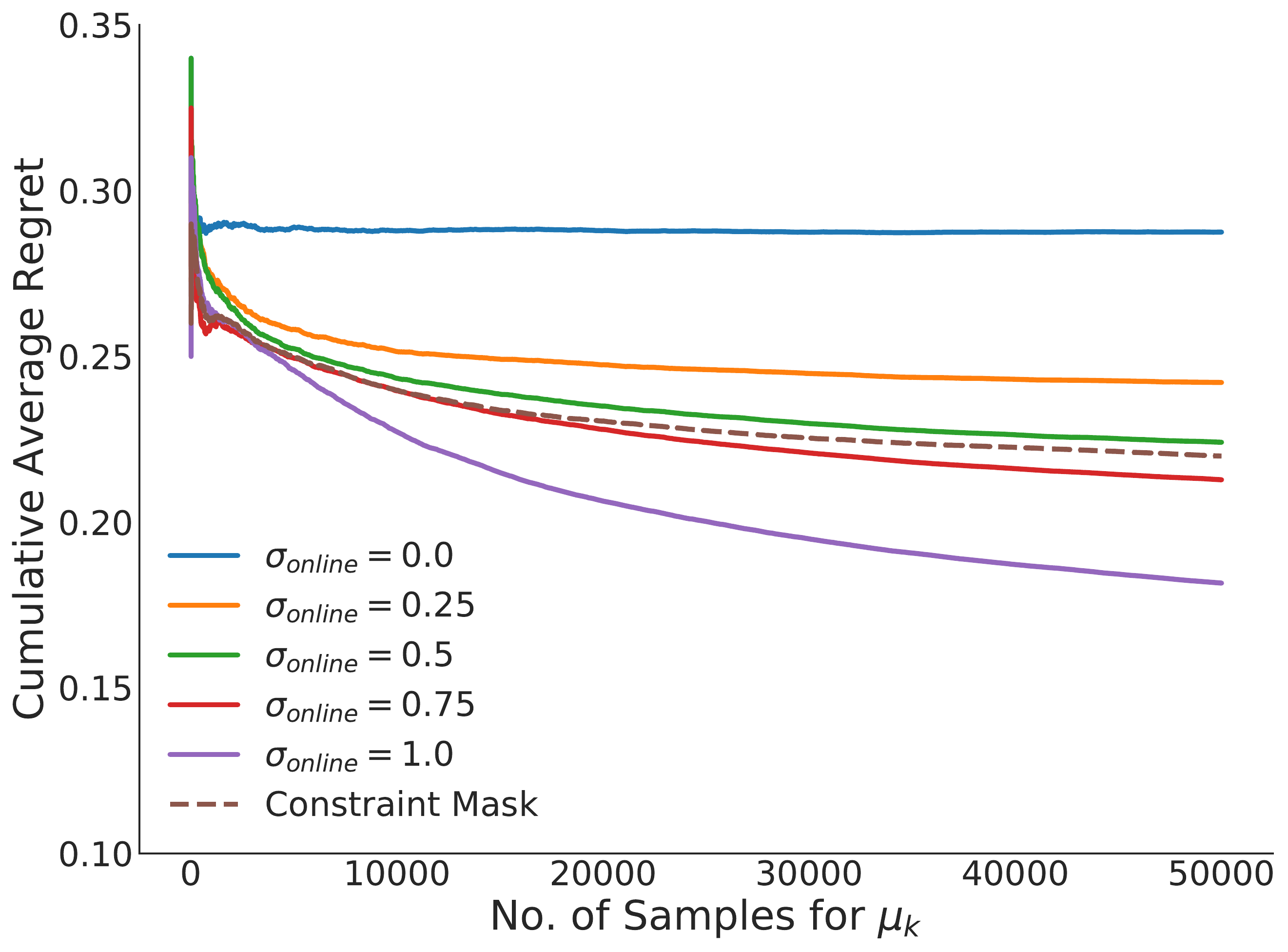}
	\caption{{\tiny Movies: $R(t)$ for CTS 50K Pretraining}}
	\label{fig:cts_movie_regret}
\end{subfigure}
\hfill
\begin{subfigure}{0.33\linewidth} 
	\centering
	\includegraphics[width=\linewidth]{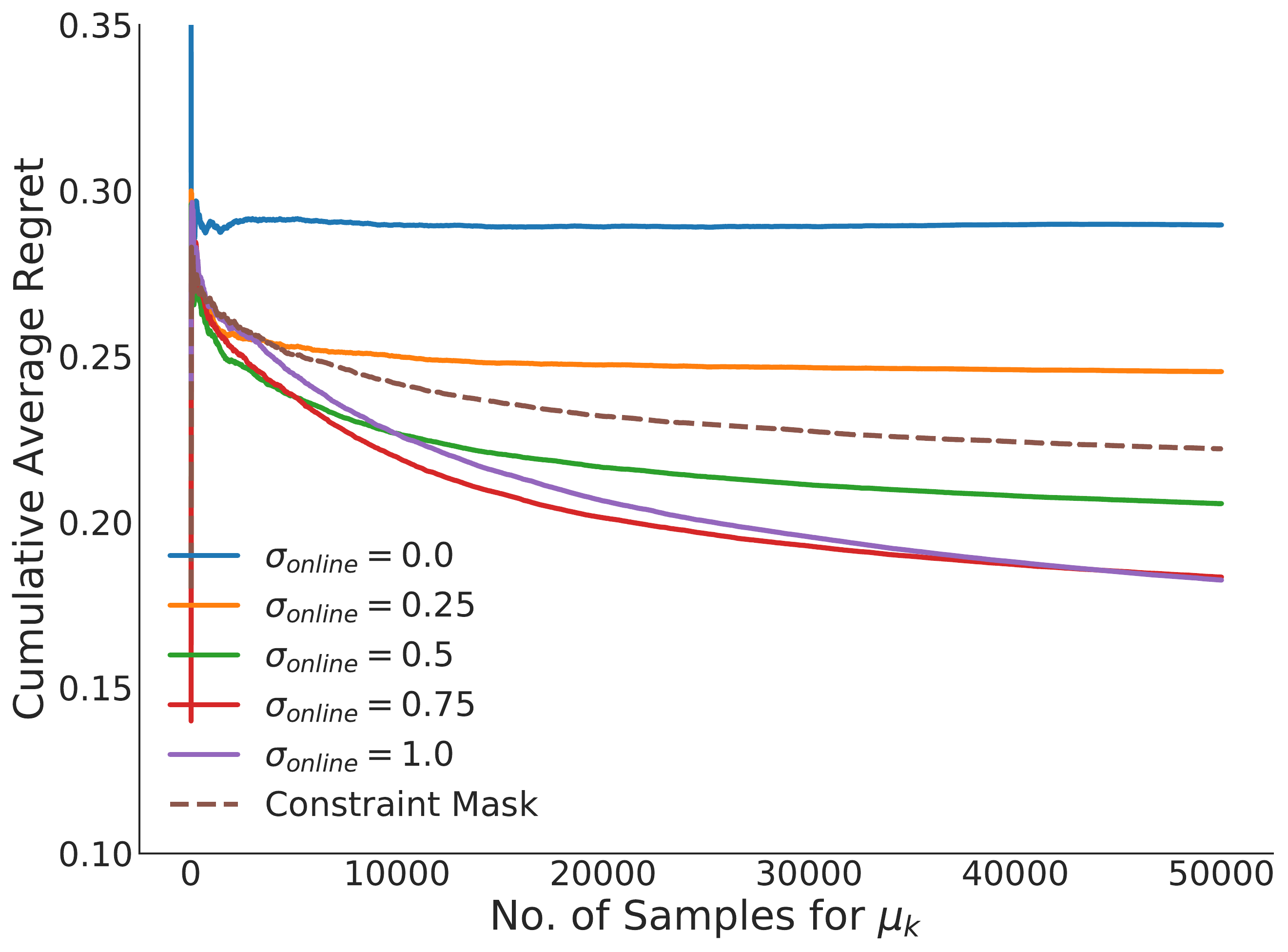}
	\caption{{\tiny Movies $R(t)$ for Random 50K Pretraining}}
	\label{fig:ran_movie_regret}
\end{subfigure}
\hfill
\begin{subfigure}{0.33\linewidth}
	\centering
	\includegraphics[width=\linewidth]{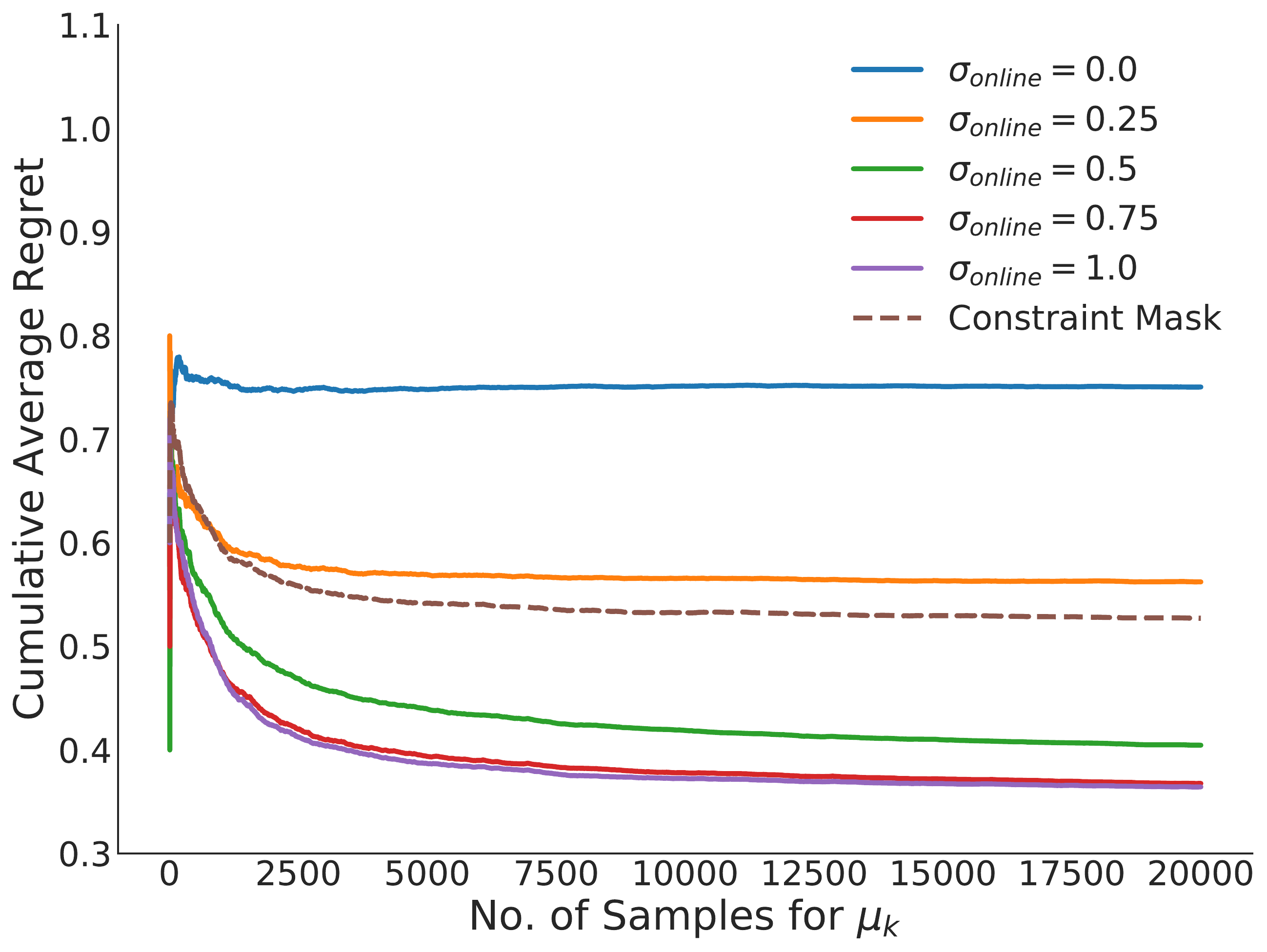}
	\caption{{\tiny Warfarin: $R(t)$ for CTS 50K Pretraining}}
	\label{fig:waf_regret}
\end{subfigure}
\begin{subfigure}{0.33\columnwidth} 
	\centering
	\includegraphics[width=\linewidth]{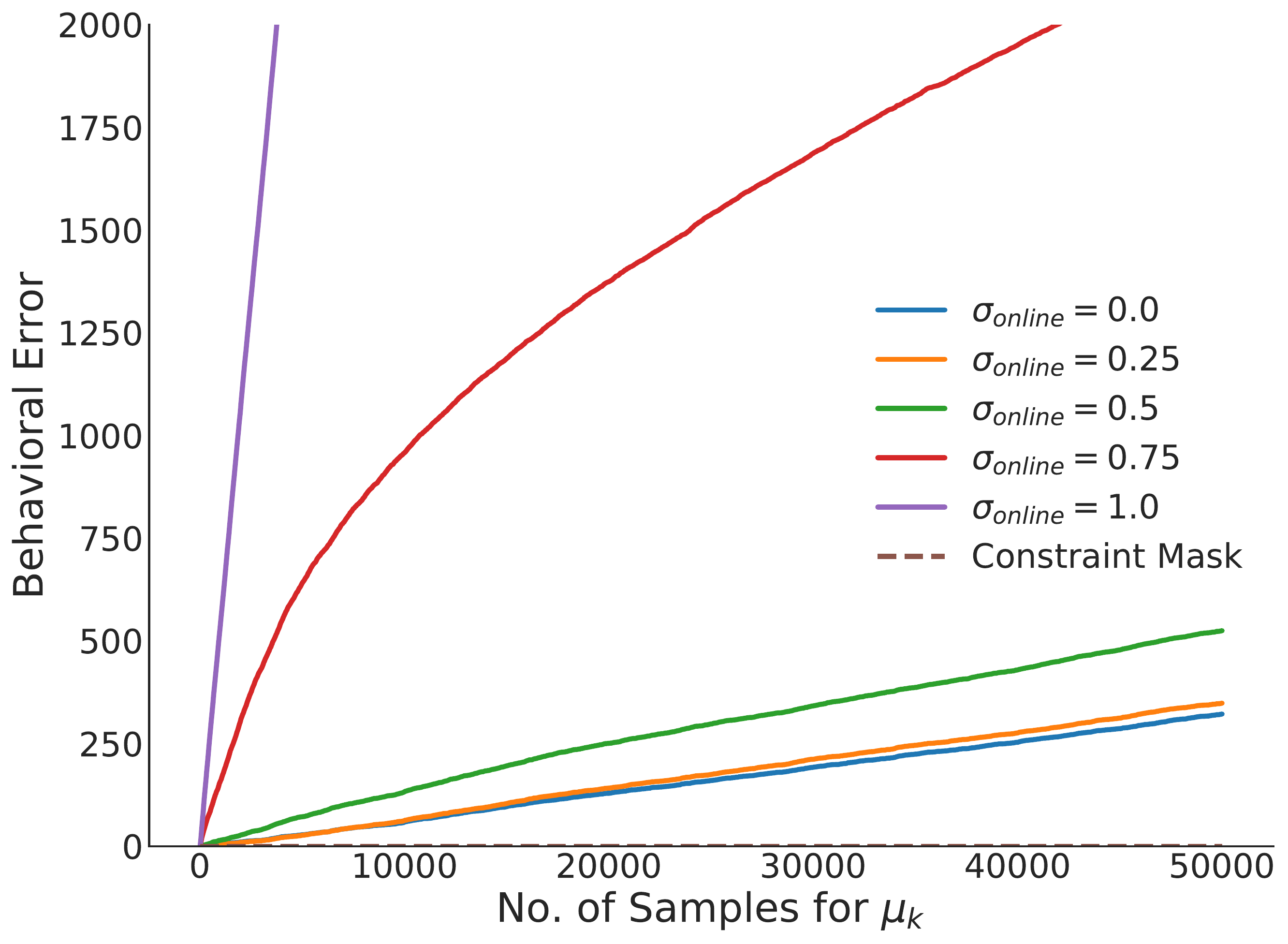}
	\caption{{\tiny Movies $E(T)$ for CTS 50K Pretraining}}
	\label{fig:cts_movie_error}
\end{subfigure}
\hfill
\begin{subfigure}{0.33\columnwidth} 
	\centering
	\includegraphics[width=\linewidth]{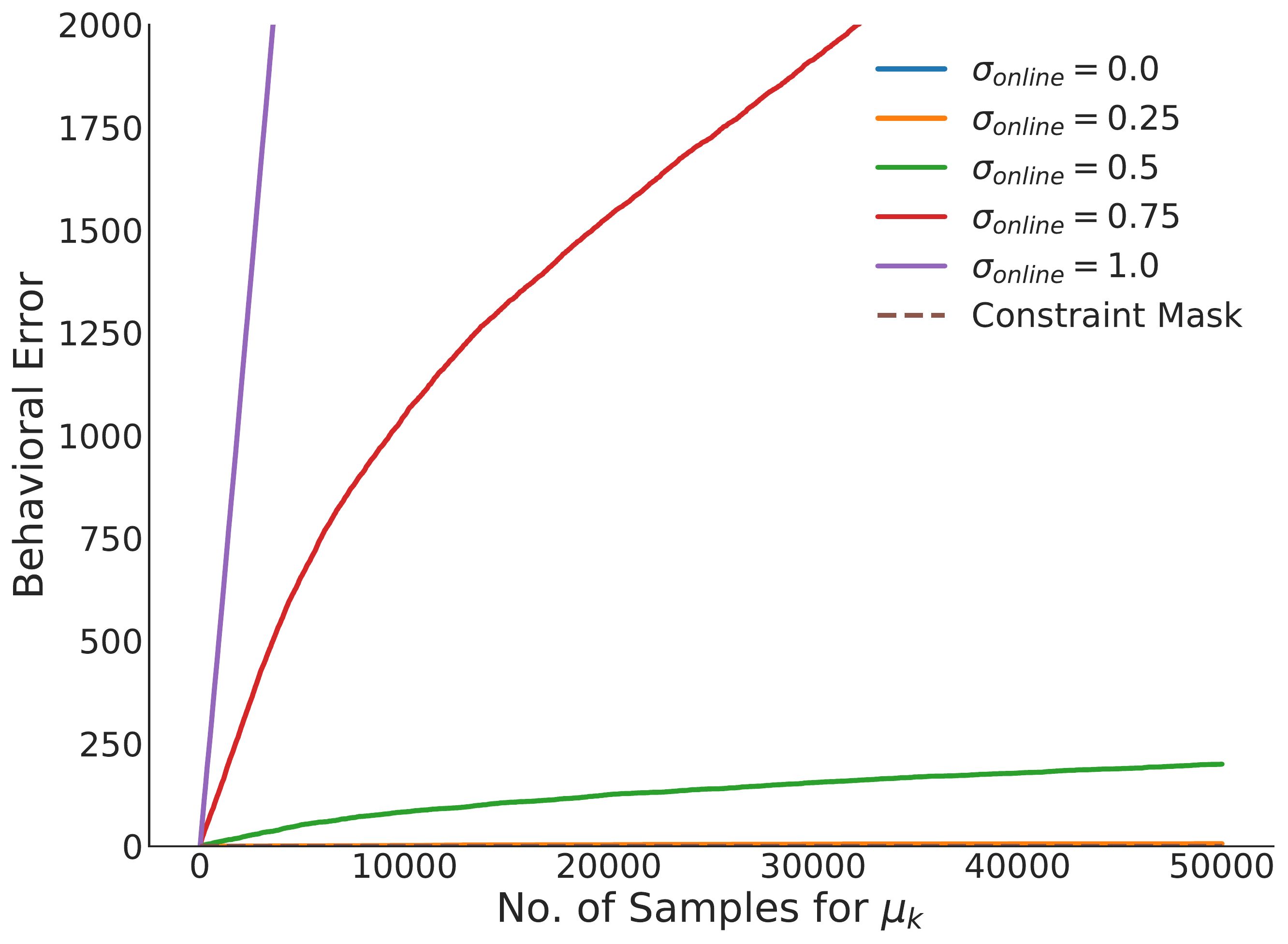}
	\caption{{\tiny Movies $E(T)$ for Random 50K Pretraining}}
	\label{fig:random_movie_error}
\end{subfigure}
\hfill
\begin{subfigure}{0.33\linewidth} 
	\centering
	\includegraphics[width=\linewidth]{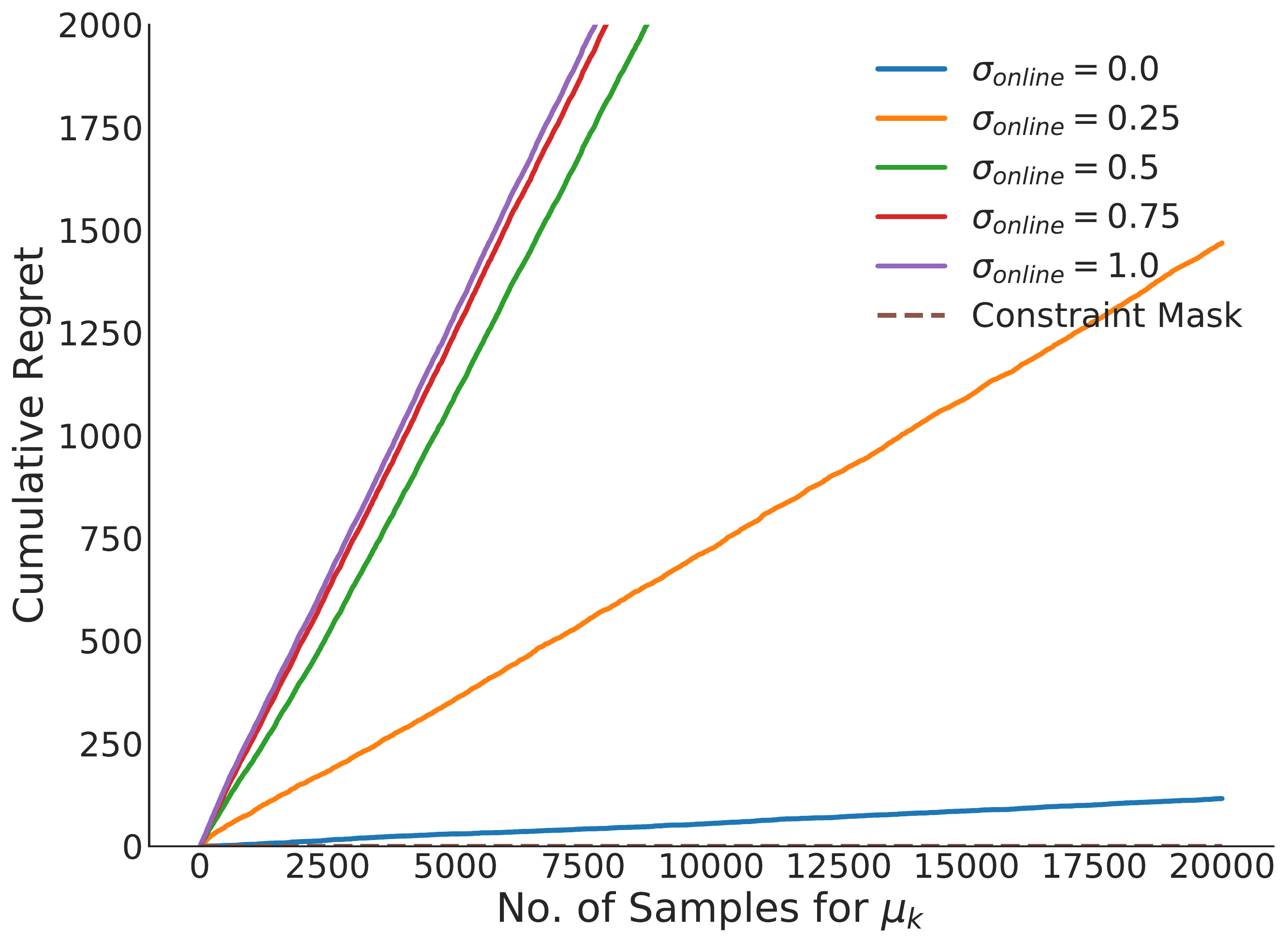}
	\caption{{\tiny Warfarin $E(t)$ for CTS 50K Pretraining}}
	\label{fig:waf_error}
\end{subfigure}

\caption{Selected results of our experiments for the movie and Warfrin datasets; mean over 5 cross validations. Figures (a--c) shows the cumulative average regret $R(t)$ as we vary $\blend$, where $\mu^e$ was trained with 50,000 examples and the Constraint Mask agent who is explicitly provided the constraints. Figures (d--f) shows the behavioral error $E(t)$, i.e., the total number of constraints violated. Increasing the sensitivity to the online reward improves the performance of the agent significantly. Notice the scale difference between the axes for the plots, $\blend=1.0$ is making a linear number of behavioral errors in both cases.  With $\blend=0.25$ for both settings we are able to make a very small number of behavioral errors within both domains and significantly reduce regret.  This implies that our agent is able to trade off $R(t)$ and $E(t)$ using the $\blend$ parameter.}
\label{fig:results}
\end{figure*}

To study the effect of imposing exogenous constraints on an online decision making agent and to demonstrate the soundness and flexibility of our techniques, we perform a set of experiments using real world data.  As we are unaware of any prior work that we can directly compare with where a constraint is learned, or inferred, and then used to bound or control the behavior of an online agent, 
we construct three reasonable baselines for comparison.  First, an agent who perfectly knows the constraints and is able to completely obey them, second, an agent who is follows a Thompson Sampling approach to learn the constraints, and finally, an agent that learns the constraints through random sampling of the space during the teaching phase.

\subsection{Movie Data}
We build an online movie recommendation agent \cite{MaryGP15} that learns online what movies to recommend to protected users (such as kids or older people) but also follows some guidelines set by parents or relatives over which movies are suitable (and which are not) for their kids (or older members of the family).  For example, parents may think that young people should not be recommended movies with too much violence, and that older people should not be recommended horror movies. 

Reward is given when the user reviews the movie, as in Netflix. We want the online agent to accrue as much reward as possible, while at the same time not violating or straying too far from the constraints.  We now describe the data and how we created the constraints in this setting.

\smallskip
\noindent
\textbf{Data.} We start from the MovieLens 20m dataset \cite{harper2016movielens}, which contains 20 million ratings of 27,000 movies by 138,000 users along with genre information. We subset this by taking the top 100 users by number of movies rated and top 1000 movies by number of ratings received. The subset of the data from MovieLens is sparse in terms of the ratings of the movies by the users. Since we resample to run many experiments over the same dataset, we create a complete rating matrix with a user based collaborative filtering pass, rounded to increments of $0.5$ to match the rest of the data \cite{BeKo07,ScFrHeSe07}. 
 
 %Formally, we are given an $n \times m$ matrix where $n$ are the rows and represent the individual users and $m$ is the columns and represent individual movies. We use the following equation for filling out the matrix for user $u$ and movie $i$. Let $\overline{r_{u}}$ denote the average rating for user $u$ on all movies and let $pred(u,i)$ denote the real value that user $u$ assigns to movie $i$ \cite{ScFrHeSe07}: $ pred(u,i) = \overline{r_u} + \frac{\sum_{n \in N} SIM(u,n)\cdot(r_{n,i} - \overline{r_n})}{\sum_{n \in N} SIM(u,n)}$. We apply a post processing rounding step so that the final rating is a multiple of $0.5$.

The genre information, which defines the context vector, for movie $m$ is $c^d(m)$ is a $d=10$ dimensional vector consisting of one or more of the categories: Action, Adventure, Comedy, Drama, Fantasy, Horror, Romance, Sci-Fi, Thriller. We impute ages to the users using the bands that marketers most often used in advertising: 12-17, 18-24, 25-34, 35-44, 45-54, 55-64, 65+ \cite{JoEl12}.  This is drawn randomly for each user using the population by age and sex demographics from the US Census.\footnote{\url{https://factfinder.census.gov/}}  

\smallskip
\noindent
\textbf{Methodology.}  The MovieLens data does not contain any notion of constraints so we must construct the constraints to test our methods.  While in the real world the we assume the behavior constraint is not given explicitly we create a \emph{behaviorally constrained training set} derived from an explicit \emph{behavior constraint matrix} to train our agent.  The behavior constraint matrix is a $\{0,1\}$ matrix of size $a \times d = 7 \times 10$ representing whether or not we can recommend a movie with genre type $d$ to someone of age type $a$.  We convert this to a behavioral training set (matrix) of size $users \times movies$ where each cell contains the behavioral reward of recommending movie $m$ to user $u$.  We enforce a constraint in the most restricted sense, i.e., if a movie has multiple features then, if any feature is restricted, the movie is as well. For example, if the behavioral constraint matrix has a $0$ in the entry at at the $(12-17) \times (\text{Horror})$ location then we are not allowed to suggest any movie $m$ which has $c(m)(\text{Horror}) = 1$ to any user (arm) $u$ which has $\textbf{f}(u)(\text{12-17})=1$. To learn the behaviorally constrained policy $\mu^e$ during the constraint learning phase we use a disjoint subset of 200 movies (contexts), but the same arms (users), from those used in the recommendation phase. During this phase we use both the classic CTS algorithm as well as random exploration and vary the number queries allowed to learn $\mu^e$. For the recommendation phase we show the contexts (movies) to the agent as we vary $\blend$ between $\blend = 0$, always follow $\mu_{e}$ and $\blend = 1$, always follow $\tilde{\mu}$.  For each setting of $\blend$ we show the movies in the same order so our results are comparable across settings of $\blend$.

\begin{figure*}
\begin{floatrow}
\capbtabbox{%
\centering
\resizebox{1.02\linewidth}{!}{
\begin{tabular}{lrr|rr|rr|rr|rr|rr}  
\toprule
& \multicolumn{2}{c}{Mask} & \multicolumn{2}{c}{$\blend=0.0$} & \multicolumn{2}{c}{$\blend=0.25$} & \multicolumn{2}{c}{$\blend=0.50$} & \multicolumn{2}{c}{$\blend=0.75$} & \multicolumn{2}{c}{$\blend=1.00$} \\
\cmidrule(r){2-13}
$N$ & $R(T)$ & $E(T)$ & $R(T)$ & $E(T)$ & $R(T)$ & $E(T)$ & $R(T)$ & $E(T)$ & $R(T)$ & $E(T)$ & $R(T)$ & $E(T)$ \\
\midrule
5K		& 0.214	& 0	& 0.310	& 787.4	& 0.279	& 572.0	& 0.240	& 878.0	& 0.204	& 4834.2	& 0.176	& 22006.8 \\
10K		& 0.216	& 0	& 0.274	& 296.0	& 0.240	& 231.8	& 0.212	& 423.2	& 0.191	& 2325.8	& 0.178	& 24706.2 \\
50K		& 0.220	& 0	& 0.288	& 322.2	& 0.242	& 348.6	& 0.224	& 525.0	& 0.212	& 2198.4	& 0.182	& 23390.4 \\
75K		& 0.213	& 0	& 0.316	& 224.2	& 0.248	& 180.0	& 0.235	& 308.4	& 0.227	& 2278.6	& 0.178	& 23611.8 \\
100K	& 0.207	& 0	& 0.284	& 489.0	& 0.220	& 542.4	& 0.206	& 734.2	& 0.201	& 2412.6	& 0.171	& 22900.0 \\
\end{tabular}
}
}{%
\caption{Performance as we vary the number of training examples $N$ for various settings of $\blend$ using CTS as the algorithm to learn the constraints with $T = 50,000$. Numbers are means over 5 cross validations.}
\label{tab:cts}
}
\capbtabbox{%
\hspace{-2em}
\centering
\resizebox{1.05\linewidth}{!}{
\begin{tabular}{lrrrrrrrrrrrr}  
\toprule
& \multicolumn{2}{c}{Mask} & \multicolumn{2}{c}{$\blend=0.0$} & \multicolumn{2}{c}{$\blend=0.25$} & \multicolumn{2}{c}{$\blend=0.50$} & \multicolumn{2}{c}{$\blend=0.75$} & \multicolumn{2}{c}{$\blend=1.00$} \\
\cmidrule(r){2-13}
$N$ & $R(T)$ & $E(T)$ & $R(T)$ & $E(T)$ & $R(T)$ & $E(T)$ & $R(T)$ & $E(T)$ & $R(T)$ & $E(T)$ & $R(T)$ & $E(T)$ \\
\midrule
5K		& 0.215	& 0	& 0.293	& 177.2	& 0.275	& 258.4	& 0.241	& 646.2	& 0.201	& 3044.8	& 0.179	& 24155.2 \\
10K		& 0.224	& 0	& 0.283	& 32.2	& 0.263	& 31.2	& 0.228	& 201.8	& 0.187	& 1986.6	& 0.183	& 27502.8 \\
50K		& 0.222	& 0	& 0.289	& 1.0	& 0.245	& 6.6	& 0.205	& 200.2	& 0.183	& 2582.4	& 0.182	& 27541.0 \\
75K		& 0.220	& 0	& 0.284	& 0.2	& 0.229	& 5.8	& 0.204	& 265.8	& 0.179	& 2433.8	& 0.181	& 27499.0 \\
100K	& 0.229	& 0	& 0.277	& 0.0	& 0.222	& 4.6	& 0.193	& 212.0	& 0.181	& 2416.6	& 0.181	& 27654.2 \\
\end{tabular}
}
}{%
\caption{Performance as we vary the number of training examples $N$ for various settings of $\blend$ when using random sampling as the algorithm to learn the constraints with $T = 50,000$. Numbers are means over 5 cross validations.}
\label{tab:random}
}
\end{floatrow}
\end{figure*}

\smallskip
\noindent
\textbf{Results.} During the online runs we track the cumulative average regret at time $T$, $R(T) = \nicefrac{\sum ^{T}_{t=1} {r_{\mu^*}(t)- r_{\tilde{\mu}}(t)}}{T}$, where  $r_{\mu^*}(t)$ is the reward for the best action the agent could have taken at time $t$ and $r_{\tilde{\mu}}$ is the reward for the action actually taken.  We also track the behavioral error at time $T$ which is given by $E(T) = \sum ^{T}_{t=1} r^e_{\tilde{\mu}}(t)$. We vary the value of $\blend$ and compare these results to an omniscient agent who knows the constraints exactly and applies them as a mask, hence it never violates the behavioral constraints and is accumulating as much reward as possible under the constraints. The results of these experiments are depicted in Figure \ref{fig:results} as well as Table \ref{tab:cts} and \ref{tab:random} where \emph{Constraint Mask} is the agent who is given the constraints explicitly. For each of these experiments, we show the means over 5 cross validations using a different subset of 200 movies to train $\mu^e$ each time. This gives us plots depicted in Figure \ref{fig:results} and the extended results in Tables \ref{tab:cts} and \ref{tab:random}.  From this we can say that the agent is consistent within this setting.

Looking first at Figures 1a and 1b we see the results when we impose a number of constraints on the agent as we set $\blend$ to various values; $\mu^e$ was trained with 50,000 samples for all graphs.  Recall that when $\blend=0.0$ the agent is not sensitive to the reward and hence has consistent behavior during the online phase, accumulating a larger amount of regret than the other agents but incurring less behavioral error. The interesting result comes when allowing some sensitivity to the online feedback ($\blend = 0.50$) as we get, for both the CTS and the random pretraining, drastically better performance in terms of balancing $R(t)$ and $E(t)$ which can be seen by contrasting with Figures \ref{fig:cts_movie_error} and \ref{fig:random_movie_error} which plot $E(t)$.  Encouragingly, looking at Figures \ref{fig:cts_movie_error} and \ref{fig:random_movie_error} we see that while the masked agent never makes any errors, the $\blend=\{0.0,0.25, 0.50\}$ agents are able to make a very small number of errors, indicating that we have been able to learn the constraints well from only the examples.  

Turning to the question of which method of pretraining is better, random or CTS, Tables \ref{tab:cts} and \ref{tab:random} give us a closer look.  These tables show $R(T)$ and $E(T)$ with $T=50,000$ for the movie domain when one agent was trained with CTS and the other with random.  Surprisingly, we see that the random pretraining agent is \emph{better} at learning the constraints across the board than the agent trained with CTS; making significantly fewer errors when $\blend \{0.00, 0.25, 0.50\}$ and decreasing the total error, $E(T)$ as we train it with more examples.  Going so far as to behave optimally when we set $\blend=0.0$ and pretrain with 100K examples.  Most interestingly, we see in Table \ref{tab:random} that the random agent for $\blend=0.25$ and 100K examples is able to both make no behavioral errors, i.e., $E(T)=0$ \emph{and} out perform the Constraint Mask in terms of $R(T)$.  From this table we can infer two things: (1) that our random agent is able to learn a high quality constrained policy given only examples and (2) that it is able to deploy this policy in the online recommendation phase.

\subsection{Healthcare Data}

Turning to a more high stakes domain, we consider a dataset for the jealthcare domain, specifically about the Warfarin drug.  Warfarin is one of the most used anticoagulants in the world and correct dosing is a standing challenge \cite{wysowski2007bleeding}.  Over 40,000 emergency room visits a year are related to Warfarin dosing \cite{budnitz2006national} and, for this reason, large datasets are available that track upwards of 50 patient factors along with the appropriate dosing levels. %The high dimensionality of this data  may cause problems for convergence in the CMAB setting, and it has been widely studied \cite{bastani2015online}.  

\smallskip
\noindent
\textbf{Data \& Methods.} In this dataset the context dimensionality $d=39$ and includes factors such as age, sex, and presence of interacting drugs including acetaminophen.  The dosing options, the arms here, is in three levels: low, medium, and high.  To create behavioral constraints we suppose that there are two additional features, randomly distributed amongst the patients, such that the presence of both of these features would cause a significant decrease in the quality of life of the patient.  We add a \emph{no dose} arm, and enforce, as a behavioral constraint, that patients should not be prescribed Warfarin if both additional features are present.  Reward in the online case is obtained by the agent for prescribing the correct dosing level, hence there will always be a cost associated with following the behavioral constraints in this setting.  The rest of our methodology is unchanged from the movie dataset.

\smallskip
\noindent
\textbf{Results.} We track the same statistics as for the movie domain. The results are depicted in Figures 1c and 1f when we train $\mu_e$ with 50,000 teaching examples. First, we see that our agent is able to learn a very good constrained policy in this space and follow it. Again, Our $\blend=0.25$ agent with 50,000 examples is able to have regret on par with the Constraint Mask agent while only making 1,500 mistakes over 20,000 trials versus nearly 20,000 for all the other agents. Allowing increasing the sensitivity to the online reward improves the performance of the agent but, in this setting, the constraints are orthogonal to the rewards. Thus the only way to collect additional reward is to violate the constraints and we see this in the high number of errors necessary in order to decrease regret.  However, the $\blend=0.25$ agent is able to perform on par with the Constraint Mask agent with only about 75 errors per 1,000 pulls. The high dimensionality the Warfarin data can cause problems for convergence when modeled as CMAB problem \cite{bastani2015online}. However, we we still achieve good results for the constrained behavior in this setting.

\subsection{Discussion}
Looking at both experimental domains, we see that using either random or CTS we are able to learn a high quality constrained policy described only by examples of appropriate behavior.  While we have theoretical bounds for the quality of our CTS agent, we see that the random agent is able to outperform it in practice. Regardless, the agent is able to use this policy to guide the actions it takes during the recommendation phase. This agent is able to make decisions that very rarely (or never) violate the constrained policy, achieving performance on par with an agent that is given the behavioral constraints explicitly. Understanding the specific interaction between the distribution of the behavioral constraints and the rewards in the data is an important direction for future work.  We saw that in some cases we are able to follow the constraints without affecting the accumulated rewards, while in other cases it had a large impact on the quality of the agents decisions. We also plan to use an active learning approach \cite{KrishnamurthyAH17} to the problem, where the online agent is presented the option to query the omniscient constraint agent during the online phase. 

\section{Conclusions}

In real life scenarios agents that recommend items to humans are subject to a plethora of exogenous priorities, given by ethical principles, moral values, social norms, and professional codes. It is essential that AI system are able to understand decisions based on their compliance to such constraints in order to reach suitable tradeoffs between satisfying user's desires and behaving appropriately. We propose to achieve this tradeoff by extending the CMAB problem machinery to include exogenous constraints, modeled in the form of a policy learned during a constraint learning phase from observing a teaching agent. Our evaluation over real data shows that our system can act within these behavior constraints while not significantly degrading the reward.  An important direction for future work is understanding the interaction between the behavioral constraints and the online rewards for various data distributions.

%An important direction for future work would be to try new scenarios, real data, and constraint functions to evaluate the performance of the online agent under a large variety of settings.  One could also use different distance functions that, hopefully, lead to a larger spread of behavior between different settings of $\blend$.  %Additionally, it would be interesting to bound the expected regret as a function of the constraints and the reward matrix.  One can imagine a case where the constraints, and hence the training examples, would be completely orthogonal to the reward in the domain.  In such a case it would be fruitful to have bounds based on the data itself.

%solve this problem through what we call "Inverse contextual bandit" where the agent don't now the reward function. However, it could learn it from an ethical person. We proposed an algorithm called "Thompson Sampling with Ethic" define an idealized formalism for an ethical learner, and conduct experiments on an online learning two toy ethical dilemmas, demonstrating the soundness and flexibility of our approach.

\bibliographystyle{aaai}
%\bibliography{bandits}

\section{Proof of Theorem 1}

\begin{proof}
 We begin by recalling some important results from the literature on contextual multi-armed bandits.

 \begin{lem}\label{fact:upperCBTS} \cite{AgrawalG13}
 With probability $1-\gamma$, where $0 \leq \gamma \leq 1$, the upper bound on the expected regret R(T) for the CTS (Algorithm \ref{alg:CTS}) in the contextual bandit problem with $K$ arms and $d$ features is:
 \begin{eqnarray*}
 %E(R(T))<
 \frac{d\gamma}{z} \sqrt{T^{z+1}} (ln (T) d) ln \frac{1}{\gamma}
 \end{eqnarray*}
 \end{lem}
 \noindent

 Using Definition 1, and line 8 of Algorithm \ref{alg:ETS}, we can write the regret at time $t$ as:
 \begin{align*}
 R(t) =  \tilde{\mu}^*_k(t)^{\top}c(t)-[\blend\tilde{\mu}_k(t)^{\top}c(t)+
 \\
 (1-\blend) \mu^e_k(t)^{\top}c(t)]
 \end{align*}
 which is equal to, 
\begin{align*}
 R(t) =  [\blend\tilde{\mu}^*_k(t)^{\top}c(t) +(1-\blend) \tilde{\mu}^*_k(t)^{\top}c(t)] 
 \\
 -[\blend\tilde{\mu}_k(t)^{\top}c(t)+(1-\blend) \mu^e_k(t)^{\top}c(t)]
 \end{align*}
 \begin{align*}
 R(t) \leq  \blend ||\tilde{\mu}^*_k(t)^{\top}c(t)-\tilde{\mu}_k(t)^{\top}c(t)||_2+\\(1-\blend) ||\tilde{\mu}^*_k(t)^{\top}c(t)-\mu^e_k(t)^{\top}c(t)||_2.
 \end{align*}

% %\nick{give some intuition of where these quantities come from? also, how are the above two equations related?  I think there needs to be an implication arrow...}
 We consider each of the two above quantities as two events: \\ $E^e(t)=||\tilde{\mu}^*_k(t)^{\top}c(t)-\mu^e_k(t)^{\top}c(t)||_2$ and $ E^o(t)=||\tilde{\mu}^*_k(t)^{\top}c(t)-\tilde{\mu}_k(t)^{\top}c(t)||_2$ .

 Starting with $E^e(t)$ and using Cauchy-Schwartz, we have $||\tilde{\mu}^*_k(t)^{\top} c(t)-\mu^{e}_k(t)^{\top}c(t) ||_2 \leq ||c(t)||_2||\tilde{\mu}^*_k(t)-\mu^{e}_k(t)||_2$. Thus the regret of $E^e(t)$ is given by:
 \begin{eqnarray*}
 	E^e(t)\leq ||c(t)||_2||\tilde{\mu}^*_k(t)-\mu^{e}_k(t)||_2.
 \end{eqnarray*}
 So, at time $T$, with $c_{max}= max_t(||c(t)||_2)$ and $t \in [1,T]$ we have,
 \begin{eqnarray*}
 	 \sum_{t=1}^{t=T} E^e(t) \leq c_{max}\sum_{t=1}^{t=T} ||\tilde{\mu}^*_k(t)-\mu^{e}_k(t)||_2.
 \end{eqnarray*}
 with $\mu^{e}_{max}= max_k(||\mu^{e}_k||_1)$, $\tilde{\mu}^*_{max}= max_k(||\tilde{\mu}^*_k||_1)$ and $k \in K$, we can write:
 \begin{eqnarray*}
 	\sum_{t=1}^{t=T} E^e(T) \leq c_{max} T ||\tilde{\mu}^*_{max}+\mu^{e}_{max}||_2.
 \end{eqnarray*}
% %
 For $E^o(t)$ we use Lemma \ref{fact:upperCBTS} to get;
 \begin{eqnarray*}
  \sum_{t=1}^{t=T} E^o(t) \leq \frac{d\gamma}{z} \sqrt{T^{z+1}} (ln (T) d) ln \frac{1}{\gamma}
 \end{eqnarray*}

 \noindent
 The two cases are independent and thus we can combine the bounds to get an upper bound for the BCTS algorithm. %$max(\blend^*,\blend)[\frac{d\gamma}{z} \sqrt{T^{z+1}} (ln (T) d) ln \frac{1}{\gamma}] + max((1-\blend^*),(1-\blend)) [D_{max} T |\mu^{e^*}_k(T)-\mu^{e}_k(T)|]$
 \end{proof}

\section{Proof of Theorem 2}

\begin{proof}
 
 Using Definition 2, and line 8 of Algorithm \ref{alg:ETS}, we can write the regret at time $t$ as:
 \begin{align*}
 R(t) =  (\blend^*\tilde{\mu}^*_{k}(t)+(1-\blend^*)\tilde{\mu}_{k}^{e*}(t))^\top c(t)-
  \\
 [\blend\tilde{\mu}_k(t)^{\top}c(t)+(1-\blend) \mu^e_k(t)^{\top}c(t)]
 \end{align*}
 which is equal to, 
 \begin{align*}
 R(t) \leq  max(\blend^*,\blend)||\tilde{\mu}^*_k(t)^{\top}c(t)-\tilde{\mu}_k(t)^{\top}c(t)||_2+\\max((1-\blend^*),(1-\blend))||\tilde{\mu}^{e*}_k(t)^{\top}c(t)-\mu^e_k(t)^{\top}c(t)||_2.
 \end{align*}

% %\nick{give some intuition of where these quantities come from? also, how are the above two equations related?  I think there needs to be an implication arrow...}
 We consider each of the two above quantities as two events: \\ $E^e(t)=||\tilde{\mu}^{e*}_k(t)^{\top}c(t)-\mu^e_k(t)^{\top}c(t)||_2$ and $ E^o(t)=||\tilde{\mu}^*_k(t)^{\top}c(t)-\tilde{\mu}_k(t)^{\top}c(t)||_2$ .

 Starting with $E^o(t)$ we use Lemma \ref{fact:upperCBTS} to get;
 \begin{eqnarray*}
  \sum_{t=1}^{t=T} E^o(t) \leq \frac{d\gamma}{z} \sqrt{T^{z+1}} (ln (T) d) ln \frac{1}{\gamma}
 \end{eqnarray*} 
 
 Also, using Lemma \ref{fact:upperCBTS} and considering that the TS was used in the training phase, the regret of $E^e(t)$ is given by: 
  \begin{eqnarray*}
  \sum_{t=1}^{t=N} E^e(t) \leq \frac{d\gamma}{z} \sqrt{N^{z+1}} (ln (N) d) ln \frac{1}{\gamma}
 \end{eqnarray*} 
% %
 \noindent
The two cases are independent and thus we can combine the bounds to get an upper bound for the BCTS algorithm.
 \end{proof}

\end{document}